\documentclass[10pt,twocolumn,letterpaper]{article}

\usepackage{iccv}
\usepackage{times}
\usepackage{epsfig}
\usepackage{graphicx}
\usepackage{amsmath}
\usepackage{amssymb}

\usepackage{algorithmic}
\usepackage{algorithm}

\usepackage{url}

\iccvfinalcopy % *** Uncomment this line for the final submission

\def\iccvPaperID{3297} % *** Enter the ICCV Paper ID here

\ificcvfinal\fi
\begin{document}

%%%%%%%%% TITLE
\title{Zero-Shot Learning with Sparse Attribute Propagation}

\author{Nanyi Fei$^{1}$~~~Jiechao Guan$^{1}$~~~Zhiwu Lu$^{1}$\thanks{Corresponding author.}~~~Tao Xiang$^{2}$~~~Ji-Rong Wen$^{1}$\\
  $^1$School of Information, Renmin University of China, Beijing 100872, China\\
  $^2$Department of Electrical and Electronic Engineering, University of Surrey, UK\\
  {\tt\small zhiwu.lu@gmail.com~~~~t.xiang@surrey.ac.uk} \\
}

\maketitle
%\thispagestyle{empty}

%%%%%%%%% ABSTRACT
\begin{abstract}
   Zero-shot learning (ZSL) aims to recognize a set of unseen classes without any training images. The standard approach to ZSL requires a set of training images annotated with seen class labels and a semantic descriptor for seen/unseen classes (attribute vector is the most widely used). Class label/attribute annotation is expensive; it thus severely limits the scalability of ZSL. In this paper, we define a new ZSL setting where only a few annotated images are collected from each seen class. This is clearly more challenging yet more realistic than the conventional ZSL setting. To overcome the resultant image-level attribute sparsity, we propose a novel inductive ZSL model termed sparse attribute propagation (SAP) by propagating attribute annotations to more unannotated images using sparse coding. This is followed by learning bidirectional projections between features and attributes for ZSL. An efficient solver is provided, together with rigorous theoretic algorithm analysis. With our SAP, we show that a ZSL training dataset can now be augmented by the abundant web images returned by image search engine, to further improve the model performance. Moreover, the general applicability of SAP is demonstrated on solving the social image annotation (SIA) problem. Extensive experiments show that our model achieves superior performance on both ZSL and SIA.
\end{abstract}

\vspace{-0.0cm}
\section{Introduction}
\vspace{-0.0cm}

Due to the difficulty of collecting sufficient training images for large-scale object recognition \cite{Russakovsky2015ImageNet,he2016cvpr,krizhevsky2012nips,simonyan2014arxiv} where convolutional neural network (CNN) is often employed, zero-shot learning (ZSL) has become topical in computer vision \cite{fu2015transductive,Fu2015CVPR,lei2015predicting,qiao2016less,ye2017cvpr,karessli2017gaze,long2017learning,Kodirov2017CVPR,Wang2018AAAI}. To recognize an unseen class without any training images, all existing ZSL models leverage a semantic space as the bridge for knowledge transfer from seen classes to unseen ones, and the semantic attribute space is most commonly used \cite{Lampert2014pami}. Given a set of seen class images, the visual features are first extracted, typically using CNN pretrained on ImageNet. With the feature representation of the images and the semantic representation of the class names, the next task is to learn a joint embedding space using the seen class data. In such a space, both feature and semantic representations are projected to be directly compared. Once the projection functions are learned, they are applied to the test images and unseen class names, and the nearest neighbour class name is found by simple search for each test image.

Although ZSL can avoid the need of collecting unseen class images for training, it still requires a large number of attribute/label annotations per seen class: hundreds of class-level attribute annotations are often needed, along with hundreds of image-level class label annotations. This severely limits the scalability of ZSL. In this paper, to study how to overcome this limitation associated with existing ZSL models and make ZSL truly scalable, we define a new ZSL setting where only a few annotated images are collected from each seen class. This is clearly more challenging yet \emph{more realistic} than the conventional ZSL setting. Note that our new ZSL setting is often encountered in real-world application scenarios such as fine-grained classification and medical image recognition. More specifically, in these scenarios, each image is hard to annotate with a class label even for an expert and thus only a few annotated images per seen class can be obtained; in the mean time, recognizing the unseen classes is always needed because the new/rare classes will unavoidably occur when more data is accumulated.

\begin{figure}[t]
\vspace{0.08in}
\centering
\includegraphics[width=0.99\columnwidth]{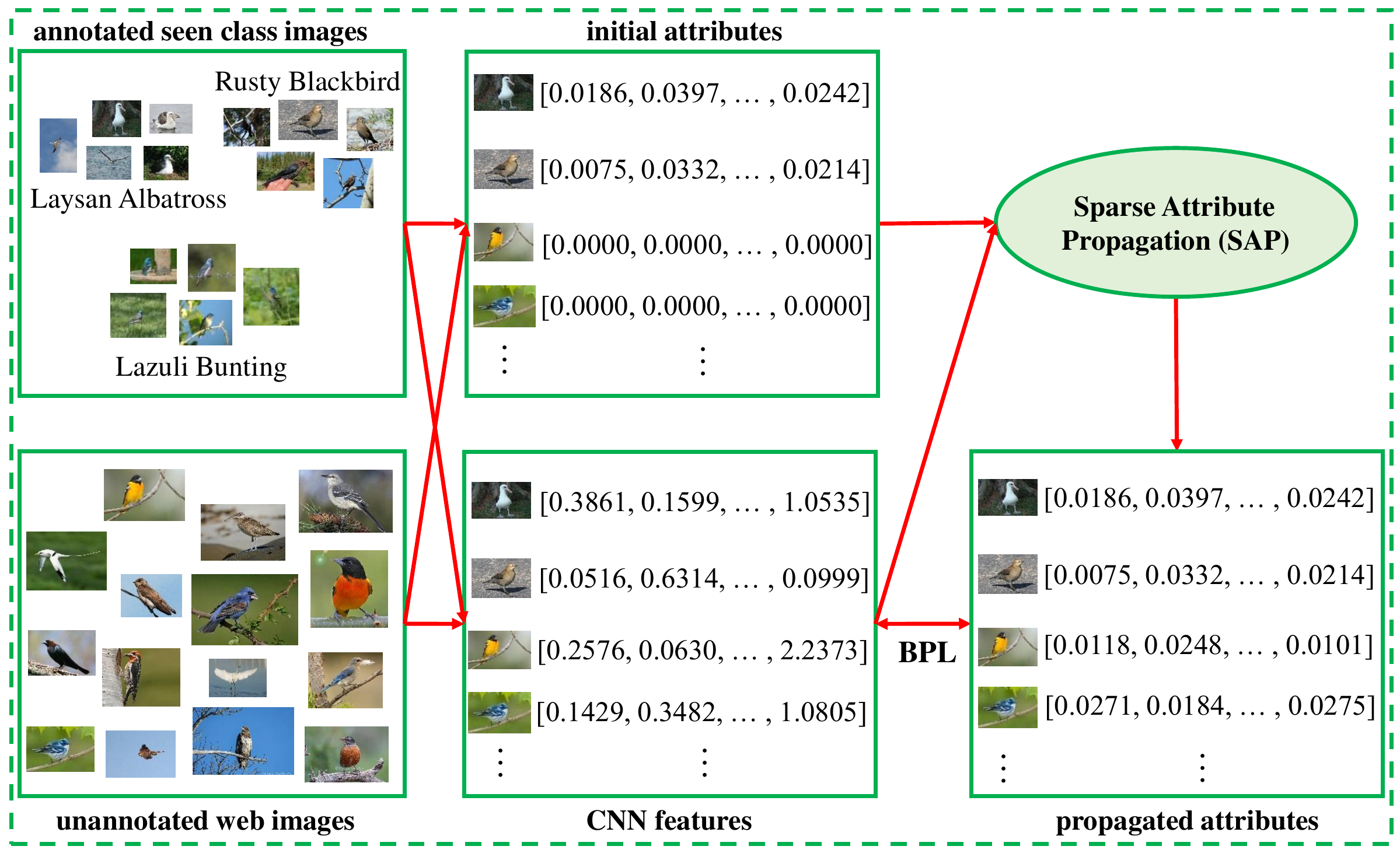}
\vspace{-0.00in}
\caption{Schematic illustration of the proposed ZSL model including SAP and BPL. The web images are obtained by Google with the query `North American Bird'. The few annotated seen class images are augmented with these unlabeled external data. }\label{fig:pipeline}
\vspace{-0.0in}
\end{figure}

To overcome the attribute sparsity under our new ZSL setting, we propose a novel inductive ZSL model termed sparse attribute propagation (SAP) by propagating attribute annotations to more unlabeled images using sparse coding \cite{LBR07,SB10}. This is followed by learning bidirectional projections between features and attributes for ZSL. We formulate sparse attribute propagation (SAP) and bidirectional projection learning (BPL) within a unified ZSL framework: SAP aims to obtain more reliable attribute annotations, while BPL aims to learn more generalizable projections. We also give an efficient iterative solver, with rigorous theoretic algorithm analysis provided. Note that under the inductive ZSL setting, only seen class images can be used for SAP. However, with SAP, our ZSL becomes a semi-supervised learning problem. As a result, we are now able to exploit the abundant web images collected using image search engine to augment a ZSL dataset. These web images could even be used to replace the unannotated seen class images which are also exploited for training. In summary, we provide a flexible ZSL approach that can scale to real-world ZSL tasks. Our ZSL model is illustrated in Figure~\ref{fig:pipeline}.

Note that our SAP model can be easily extended to other partially labelled image classification tasks. To demonstrate that, we choose the social image annotation (SIA) task \cite{johnson2015iccv,wang2016cvpr,jin2016icpr,hu2016cvpr,liu2017semantic}. In this task, the social tags associated with each image are used to form the semantic space. These user-provided social tags are rather noisy and sparse, offering only a partial semantic description of the image content. With the proposed SAP model, the SIA problem can be solved as follows: (1) The noise in social tags is reduced by sparse coding; (2) The noise reduction problem is further regularized by BPL to obtain better results.

Our contributions are: (1) For the first time, we define a new ZSL setting with only a few annotated seen class images, which is more challenging yet more realistic than the conventional ZSL setting. (2) To overcome the attribute sparsity under our new setting, we propose a novel inductive ZSL model by integrating SAP and BPL into a unified framework. An efficient iterative algorithm is formulated, together with rigorous theoretic algorithm analysis. (3) The proposed algorithm is highly flexible and can be generalized to other vision problems such as SIA. Extensive experiments show that our model is superior to the state-of-the-art alternatives on both problems (i.e. ZSL and SIA).

\section{Related Work}

\noindent\textbf{ZSL with Less Human Annotation.} A ZSL model typically exploits two types of human annotation for recognizing unseen classes without any training images: (1) the human-annotated class labels of training images from seen classes; (2) the human-defined semantic representations of seen/unseen classes. In the area of ZSL, much attention has been paid to reducing the annotation cost of generating human-defined semantic representations, i.e., the semantic space is formed using online textual documents \cite{lei2015predicting,qiao2016less}, human gaze \cite{karessli2017gaze}, or visual similes \cite{long2017describing,long2017learning} (instead of attributes), which leads to significantly less annotation cost. Different from these ZSL models, we focus on ZSL with less human annotation by defining a new ZSL setting, i.e., only a few annotated images are collected from each seen class. Although our ZSL model is proposed based on attributes in this paper, it can be easily generalized to other forms of semantic space \cite{lei2015predicting,karessli2017gaze,long2017learning,qiao2016less} to further reduce the annotation cost. To our best knowledge, we are the first to define this new ZSL setting with only a few annotated seen class images in the area of ZSL.

\vspace{0.05cm}

\noindent\textbf{Semi-Supervised ZSL.} In this paper, attribute propagation is performed from a few annotated seen class images to more unannotated images so that more reliable attribute annotations can be obtained. This can be regarded as a form of semi-supervised ZSL. Note that the test images from unseen classes are not used for training our model, i.e., we take an inductive ZSL setting. However, in the area of ZSL, when semi-supervised learning is applied to ZSL, the unlabelled test images from unseen classes are typically combined with the labelled training images from seen classes to form the training set. This results in a transductive ZSL setting: either label propagation \cite{fu2015transductive,Fu2015CVPR,li2017tgrs,ye2017cvpr} or self-training \cite{li2015iccv,guo2016aaai,shojaee2016semi,wang2017ijcv,yu2017transductive,Zhao2018nips} is employed for semi-supervised learning. Since these transductive ZSL models assume the access to the whole test set, they have limited applications in real-world scenarios. Note that although the test set is not involved in the training process, our model still exploits the unannotated seen class images for attribute propagation. Given that it is not easy to obtain the unannotated seen class images, we choose to perform attribute propagation with unannotated external data from image search engine, which thus provides a feasible/convenient approach to applying our model to real-world ZSL tasks.

\vspace{0.05cm}

\noindent\textbf{ZSL with Web Images.} In computer vision, web images have been widely used to promote the performance of existing recognition models as in \cite{bergamo2010exploiting,duan2012exploiting,NTVS2018CVPR,ZLZ2016CVPR}. However, there is less attention on exploiting web images for ZSL. Two exceptions are: the web images are utilized to augment the unseen class data in \cite{niu2018webly} and discover event composition knowledge for zero-shot event detection in \cite{gan2017deck}. In this work, although web images are also employed as external data, our model is quite different from \cite{niu2018webly} in that we do not search web images \emph{directly with unseen class names} since this is against the zero-shot setting.

\vspace{0.05cm}

\noindent\textbf{Social Image Annotation.} A number of recent works on image annotation attempted to improve the performance by exploiting side information collected from social media websites, which is called social image annotation (SIA) in this paper. The user-provided side information can be extracted from the noisy tags \cite{johnson2015iccv,liu2017semantic} and group labels \cite{hu2016cvpr}. By forming the semantic space using the social tags, our algorithm originally developed for ZSL can also be generalized to SIA. Although the label correlation is not considered in our model, it is shown to generally outperform the state-of-the-art alternatives \cite{wang2016cvpr,jin2016icpr,hu2016cvpr,liu2017semantic} that employ the well-known recurrent neural network (RNN) \cite{Graves2009A} to model the label correlation for SIA (see Table~\ref{tab:sia}).

\section{Methodology}

\subsection{Problem Definition}

Let $\mathcal{C}_s=\{cs_1,...,cs_p\}$ denote a set of seen classes and $\mathcal{C}_u=\{cu_1,...,cu_q\}$ denote a set of unseen classes, where $p$ and $q$ are the numbers of seen and unseen classes, respectively. These two sets of classes are disjoint. Similarly, $\mathbf{Z}_s = [\mathbf{z}_1^{(s)},..., \mathbf{z}_p^{(s)}] \in \mathbb{R}^{k \times p}$ and $\mathbf{Z}_u = [\mathbf{z}_1^{(u)},...,\mathbf{z}_q^{(u)}] \in \mathbb{R}^{k \times q}$ denote the corresponding seen and unseen class semantic representations (e.g. $k$-dimensional attribute vector). We are given a set of seen class training images $\mathcal{D}_s=\{(\mathbf{x}_i^{(s)}, l_i^{(s)}), \mathbf{y}_i^{(s)}: i=1,...,r,r+1,...,N_s\}$, where $\mathbf{x}_i^{(s)} \in \mathbb{R}^{d\times1}$ is the $d$-dimensional feature vector of the $i$-th training image, $l_i^{(s)} \in \{1,...,p\}$ is the label of $\mathbf{x}_i^{(s)}$ according to $\mathcal{C}_s$, $\mathbf{y}_i^{(s)}=\mathbf{z}_{l_i^{(s)}}^{(s)}$  is the semantic representation of $\mathbf{x}_i^{(s)}$ (i.e. \emph{only class-level attributes are needed}), and $N_s$ denotes the number of training images. In this paper, only the first $r$ annotated training images $\mathbf{x}_i^{(s)} (1 \le i \le r)$ have non-zero attribute vectors, i.e. $\mathbf{y}_i^{(s)} = \mathbf{0}~(r+1 \le i \le N_s)$. Moreover, let $\mathcal{D}_u= \{(\mathbf{x}_i^{(u)}, l_i^{(u)}), \mathbf{y}_i^{(u)}: i=1,...,N_u\}$ denote a set of unseen class test images, where $\mathbf{x}_i^{(u)} \in \mathbb{R}^{d\times1}$ is the $d$-dimensional feature vector of the $i$-th test image, $l_i^{(u)} \in \{1,...,q\}$ is the unknown label of $\mathbf{x}_i^{(u)}$ according to $\mathcal{C}_u$, $\mathbf{y}_i^{(u)}=\mathbf{z}_{l_i^{(u)}}^{(u)}$ is the unknown semantic representation of $\mathbf{x}_i^{(u)}$, and $N_u$ denotes the number of test images. The goal of zero-shot learning is to predict the labels of test images by learning a classifier $f:\mathcal{X}_u \rightarrow \mathcal{C}_u$, where $\mathcal{X}_u = \{\mathbf{x}_i^{(u)}: i=1,...,N_u\}$. In a generalized ZSL setting \cite{chao2016empirical,rahman2017unified,Xian2017CVPR}, the test samples can come from both seen and unseen classes, so the classifier becomes $f:\mathcal{X}\rightarrow \mathcal{C}_s \cup \mathcal{C}_u$, where $\mathcal{X}$ denotes the set of all test samples.

Note that the above problem definition is strictly consistent with the inductive ZSL setting, where the unannotated seen class training images are given along with the annotated seen class training ones. Their uniform notations make the following model formulation more concise. As we have mentioned, the unannotated seen class images can be readily replaced by the unannotated web images from image search engine (see Figure~\ref{fig:pipeline}), resulting in a feasible approach for real-world ZSL tasks. The details of ZSL with external data are given at the end of this section.

\subsection{Model Formulation}

When learned with only a few annotated images per seen class under our new ZSL setting, the projection function between the feature and semantic spaces is not reliable. Therefore, we choose to propagate such sparse attribute annotations to more unannotated images using sparse coding \cite{LBR07,SB10}: more attribute annotations enable us to learn a more reliable projection, but the noise caused by attribute propagation should also be suppressed by sparse coding, which is thus called sparse attribute propagation (SAP). Moreover, given all seen class training images (with ground truth/predicted attribute vectors), we integrate the forward and reverse projections for ZSL, since either projection suffers from the projection domain shift \cite{fu2015transductive,Kodirov2015ICCV}. By bidirectional projection learning (BPL), a visual feature vector is first projected into a semantic space and then back into visual feature space to reconstruct itself. Such self-reconstruction can improve model generalization ability and help tackle the projection domain shift. Our unified framework including SAP and BPL is given below.

Specifically, with the whole seen class training set $\mathcal{X}_s = \{\mathbf{x}_i^{(s)}: i=1,...,N_s\}$, we construct first a graph $\mathcal{G} = \{\mathcal{V}, \mathbf{A}\}$ with its vertex set $\mathcal{V} = \mathcal{X}_s$ and affinity matrix $\mathbf{A} = [a_{ij}]_{N_s \times N_s}$, where $a_{ij}$ denotes the similarity between training images $\mathbf{x}_i^{(s)}$ and $\mathbf{x}_j^{(s)}$. The affinity matrix $\mathbf{A}$ can be defined as: $a_{ij} = \exp(- \| \mathbf{x}_i^{(s)} - \mathbf{x}_j^{(s)} \| _2^2 / (2\sigma^2))$, where the parameter $\sigma$ can be determined empirically ($\sigma = 1$ in this paper). The normalized Laplacian matrix $\mathbf{L}$ is given by
\begin{small}
\begin{align}
\mathbf{L} = \mathbf{I} - \mathbf{D}^{-\frac{1}{2}} \mathbf{A} \mathbf{D}^{-\frac{1}{2}},
\label{eq:laplace}
\end{align}
\end{small}
\hspace{-3pt}where $\mathbf{I}$ is an $N_s \times N_s$ identity matrix, and $\mathbf{D}$ is an $N_s \times N_s$ diagonal matrix with its $i$-th diagonal element being $\sum_j a_{ij}$. We derive a new matrix $\mathbf{B} \in \mathbb{R}^{N_s \times N_s}$ from $\mathbf{L}$:
\begin{small}
\begin{align}
\mathbf{B} = \mathbf{\Sigma}^{\frac{1}{2}} \mathbf{V}^{T},
\end{align}
\end{small}
\hspace{-3pt}where $\mathbf{V}$ is an $N_s \times N_s$ orthonormal matrix with each column being an eigenvector of $\mathbf{L}$, and $\mathbf{\Sigma}$ is an $N_s \times N_s$ diagonal matrix with its diagonal element $\Sigma_{ii}$ being an eigenvalue of $\mathbf{L}$ (sorted as $0 \le \Sigma_{11} \le \cdots \le \Sigma_{N_sN_s}$). Denoting the eigen-decomposition of $\mathbf{L}$ as $\mathbf{L} = \mathbf{V} \mathbf{\Sigma} \mathbf{V}^T$, $\mathbf{L}$ can be represented as: $\mathbf{L} = (\mathbf{\Sigma}^{\frac{1}{2}} \mathbf{V}^T)^T \mathbf{\Sigma}^{\frac{1}{2}} \mathbf{V}^T = \mathbf{B}^T \mathbf{B}$.

We further collect the feature and attribute vectors of the training set as $\mathbf{X}^{(s)} = [\mathbf{x}_1^{(s)},...,\mathbf{x}_{N_s}^{(s)}] \in \mathbb{R}^{d \times N_s}$ and $\mathbf{Y}^{(s)} = [\mathbf{y}_1^{(s)},...,\mathbf{y}_{N_s}^{(s)}] \in \mathbb{R}^{k \times N_s}$. Our ZSL model solves the following optimization problem:
\begin{small}
\begin{align}
& \hspace{-0.05in}\min_{\mathbf{Y},\tilde{\mathbf{Y}},\mathbf{W}} \{\|\mathbf{Y} \hspace{-0.05in}-\hspace{-0.04in} \tilde{\mathbf{Y}}\|^2_F + \lambda_1 \|\mathbf{B} \tilde{\mathbf{Y}}^T\|_1 + \lambda_2 \|\mathbf{Y} \hspace{-0.05in}-\hspace{-0.04in} \mathbf{Y}^{(s)}\|_1 \nonumber \\
& \hspace{0.05in}+\hspace{-0.04in}\lambda_3 (\|\mathbf{W} \mathbf{X}^{(s)} \hspace{-0.05in}-\hspace{-0.04in} \mathbf{Y}\|^2_F \hspace{-0.03in}+\hspace{-0.03in} \|\mathbf{X}^{(s)} \hspace{-0.05in}-\hspace{-0.04in} \mathbf{W}^T \mathbf{Y}\|^2_F\hspace{-0.03in}+\hspace{-0.03in}\lambda_4\|\mathbf{W}\|^2_F) \},
\label{eq:obj}
\end{align}
\end{small}
\hspace{-3pt}where $\mathbf{W} \in \mathbb{R}^{k \times d}$ is a projection matrix from the visual feature space to the semantic space, $\mathbf{Y} \in \mathbb{R}^{k \times N_s}$ collects the optimal attribute vectors of all seen class training images, and $\tilde{\mathbf{Y}} \in \mathbb{R}^{k \times N_s}$ denotes an intermediate matrix that approaches $\mathbf{Y}$, and $\lambda_1$, $\lambda_2$, $\lambda_3$, $\lambda_4$ are free parameters.

The first and third terms of the objective function are the $L_2$-norm and $L_1$-norm fitting constraints, respectively. Particularly, the third term enforces the noise sparsity in $\mathbf{Y}$, which is a commonly used constraint for data noise and has been proven to be effective. The second term is a graph smoothness constraint, different from the conventional graph smoothness constraint as a trace norm term. Here, $L_1$-norm is used to promote the sparsity on the inferred attribute vectors and thus noise reduction (see Figure~\ref{fig:sparse}). Additionally, the last three terms denote the loss function of projection learning for ZSL. The two projection matrices are transpose of each other, similar to those in an autoencoder \cite{feng2014cross,lu2013speech}.

\subsection{Optimization Algorithm}
\label{sec:alg}

Let $\mathcal{F}(\mathbf{Y},\tilde{\mathbf{Y}},\mathbf{W})$ denote the objective function in Eq.~(\ref{eq:obj}). The optimization problem in Eq.~(\ref{eq:obj}) can be solved in two alternating steps as follows:
\begin{small}
\begin{align}
& \mathbf{SAP:}~\mathbf{Y}^*,\tilde{\mathbf{Y}}^* = \arg\min_{\mathbf{Y}, \tilde{\mathbf{Y}}} \mathcal{F}(\mathbf{Y}, \tilde{\mathbf{Y}}, \mathbf{W}^*), \label{eq:apobj} \\
& \mathbf{BPL:}~\mathbf{W}^* = \arg\min_{\mathbf{W}} \mathcal{F}(\mathbf{Y}^*, \tilde{\mathbf{Y}}^*, \mathbf{W}), \label{eq:plobj}
\end{align}
\end{small}
\hspace{-2.5pt}where $\mathbf{Y}^*$ is initialized with $\mathbf{Y}^{(s)}$, and $\mathbf{W}^*$ is initialized by solving the BPL problem in Eq.~(\ref{eq:plobj}) with $\mathbf{Y}^* = \tilde{\mathbf{Y}}^* = \mathbf{Y}^{(s)}$.

\vbox{}\noindent\textbf{Sparse Attribute Propagation (SAP).} The SAP subproblem in Eq.~(\ref{eq:apobj}) is solved with the alternating optimization technique as follows: 1) SAP-I: fix $\mathbf{Y}=\mathbf{Y}^*$, and update $\tilde{\mathbf{Y}}$ by $\tilde{\mathbf{Y}}^* = \arg\min_{\tilde{\mathbf{Y}}} \mathcal{F}(\mathbf{Y}^*, \tilde{\mathbf{Y}}, \mathbf{W}^*)$; 2) SAP-II: fix $\tilde{\mathbf{Y}}=\tilde{\mathbf{Y}}^*$, and update $\mathbf{Y}$ by $\mathbf{Y}^* = \arg\min_{\mathbf{Y}} \mathcal{F}(\mathbf{Y}, \tilde{\mathbf{Y}}^*, \mathbf{W}^*)$.

\noindent\textbf{1) SAP-I.} Directly solving the SAP-I subproblem is of high computational cost mainly due to the dimension of $\mathbf{B}$ ($N_s \times N_s$). Fortunately, we find a way to dramatically reduce this dimension by using only a small subset of eigenvectors of $\mathbf{L}$. Specifically, we decompose $\tilde{\mathbf{Y}}$ to $\tilde{\mathbf{Y}} = (\mathbf{V}_m \alpha)^T$, where $\alpha = \{\alpha_{ij}\}_{m \times k}$ is an $m \times k$ matrix that collects the reconstruction coefficients and $\mathbf{V}_m$ is an $N_s \times m$ matrix whose columns are the $m$ smallest eigenvectors of $\mathbf{L}$ (i.e. the first $m$ columns of $\mathbf{V}$). The SAP-I subproblem can be reformulated as follows:
\begin{small}
\begin{align}
\alpha^* = & \arg\min_{\alpha} \|\mathbf{V}_m \alpha - \mathbf{Y}^{*T}\|_F^2 + \lambda_1 \|\mathbf{B} \mathbf{V}_m \alpha\|_1 \nonumber \\
= & \arg\min_{\alpha} \sum_{j=1}^k (\|\mathbf{V}_m \alpha_{.j} - \mathbf{Y}^{*T}_{.j}\|_2^2 + \lambda_1 \|\mathbf{B} \mathbf{V}_m \alpha_{.j}\|_1),
\label{eq:apobj2}
\end{align}
\end{small}
\hspace{-3pt}where $\alpha_{.j}$ and $\mathbf{Y}^{*T}_{.j}$ denote the $j$-th column of $\alpha$ and $\mathbf{Y}^{*T}$, respectively. The above problem can be further decomposed into the following $k$ independent subproblems:
\begin{small}
\begin{align}
& \arg\min_{\alpha_{.j}} \|\mathbf{V}_m \alpha_{.j} - \mathbf{Y}^{*T}_{.j}\|_2^2 + \lambda_1 \|\mathbf{B} \mathbf{V}_m \alpha_{.j}\|_1 \nonumber \\
= & \arg\min_{\alpha_{.j}} \|\mathbf{V}_m \alpha_{.j} - \mathbf{Y}^{*T}_{.j}\|_2^2 + \lambda_1 \|\sum_{i=1}^m \mathbf{\Sigma}^{\frac{1}{2}} \mathbf{V}^T \mathbf{V}_{.i} \alpha_{ij}\|_1 \nonumber \\
= & \arg\min_{\alpha_{.j}} \|\mathbf{V}_m \alpha_{.j} - \mathbf{Y}^{*T}_{.j}\|_2^2 + \lambda_1 \sum_{i=1}^m \Sigma_{ii}^{\frac{1}{2}} |\alpha_{ij}|,
\label{eq:apobj3}
\end{align}
\end{small}
\hspace{-3pt}where the orthonormality of $\mathbf{V}$ is used to simplify the $L_1$-norm term $\|\mathbf{B} \mathbf{V}_m \alpha_{.j}\|_1$. Many off-the-shelf solvers exist for solving $L_1$-optimization problems like Eq.~(\ref{eq:apobj3}). The L1General toolbox\footnote{\url{https://www.cs.ubc.ca/~schmidtm/Software/L1General.html}} is employed in this paper, which can solve Eq.~(\ref{eq:apobj3}) at a linear time cost.

To further improve the efficiency, we compute the affinity matrix $\mathbf{A}$ over a $k_g$-nearest neighbor graph with $k_g \ll N_s$. The time complexity for finding $m$ eigenvectors with the smallest eigenvalues of the sparse matrix $\mathbf{L}$ is $O(m^3 + m^2N_s + k_gmN_s)$, which scales well to the data.

\noindent\textbf{2) SAP-II.} Let $\mathbf{\bar{Y}} = \mathbf{Y} - \mathbf{Y}^{(s)}$. The SAP-II subproblem can be reformulated as follows:
\begin{small}
\begin{align}
& \hspace{0.05in}\mathbf{\bar{Y}}^* = \arg\min_{\mathbf{\bar{Y}}} \|\mathbf{\bar{Y}} + \mathbf{Y}^{(s)} - \tilde{\mathbf{Y}}^*\|^2_F + \lambda_2 \|\mathbf{\bar{Y}}\|_1 \nonumber \\
& \hspace{0.25in} + \hspace{-0.03in} \lambda_3 (\|\mathbf{W}^* \mathbf{X}^{(s)} \hspace{-0.05in}-\hspace{-0.04in} (\mathbf{\bar{Y}} \hspace{-0.05in}+\hspace{-0.04in} \mathbf{Y}^{(s)})\|^2_F \hspace{-0.03in} +\hspace{-0.03in} \|\mathbf{X}^{(s)} \hspace{-0.05in}-\hspace{-0.04in} \mathbf{W}^{*T} \hspace{-0.04in} (\mathbf{\bar{Y}} \hspace{-0.05in}+\hspace{-0.04in} \mathbf{Y}^{(s)})\|^2_F)\nonumber \\
& \hspace{0.22in} = \arg\min_{\mathbf{\bar{Y}}} \mathrm{loss}(\mathbf{\bar{Y}})+ \lambda_2 \|\mathbf{\bar{Y}}\|_1,
\label{eq:scobj2}
\end{align}
\end{small}
\hspace{-4pt}where $\mathrm{loss}(\mathbf{\bar{Y}})= \|\mathbf{\bar{Y}} + \mathbf{Y}^{(s)} - \tilde{\mathbf{Y}}^*\|^2_F + \hspace{-0.03in} \lambda_3 (\|\mathbf{W}^* \mathbf{X}^{(s)} \hspace{-0.05in}-\hspace{-0.04in} (\mathbf{\bar{Y}} \hspace{-0.05in}+\hspace{-0.04in} \mathbf{Y}^{(s)})\|^2_F \hspace{-0.03in} +\hspace{-0.03in} \|\mathbf{X}^{(s)} \hspace{-0.05in}-\hspace{-0.04in} \mathbf{W}^{*T} \hspace{-0.04in} (\mathbf{\bar{Y}} \hspace{-0.05in}+\hspace{-0.04in} \mathbf{Y}^{(s)})\|^2_F)$. Since $\mathrm{loss}(\mathbf{\bar{Y}})$ is a quadratic function with respect to $\mathbf{\bar{Y}}$, the above $L_1$-optimization problem can also be solved using the L1General toolbox at a linear time cost.

\vbox{}\noindent\textbf{Bidirectional Projection Learning (BPL).} By setting $\frac{\partial \mathcal{F}(\mathbf{Y}^*, \tilde{\mathbf{Y}}^*, \mathbf{W})}{\partial \mathbf{W}} = 0$, the solution to the BPL subproblem in Eq.~(\ref{eq:plobj}) can be found by solving a Sylvester equation:
\begin{small}
\begin{align}
(\mathbf{Y}^* \mathbf{Y}^{*T}+{\lambda_4}\mathbf{I}) \mathbf{W} + \mathbf{W} (\mathbf{X}^{(s)} \mathbf{X}^{(s)T}) = 2 \mathbf{Y}^* \mathbf{X}^{(s)T},
\label{eq:plobj2}
\end{align}
\end{small}
\hspace{-3pt}which can be solved (using Matlab built-in function) with a time complexity of $O((k^2+d^2+kd)N_s+k^3+d^3)$. In this paper, we empirically set $\lambda_4=0.01$ in all experiments.

\begin{algorithm}[t]
\caption{Inductive ZSL with Joint SAP and BPL}
\label{alg:ours}
\begin{algorithmic}
\STATE {\bfseries Input:} Feature representation of the training set $\mathbf{X}^{(s)}$\\
\qquad~~~~Initial semantic representation $\mathbf{Y}^{(s)}$\\
\qquad~~~~Parameters $k_g, m, \lambda_1, \lambda_2, \lambda_3$
\STATE {\bfseries Output:} $\mathbf{W}^*$
\STATE 1. Construct a $k_g$-nearest neighbor graph with its affinity matrix $\mathbf{A}$ being defined over $\mathbf{X}^{(s)}$;\\
\STATE 2. Find $m$ smallest eigenvectors of the Laplacian matrix $\mathbf{L}$ and store them in $\mathbf{V}_m$;\\
\STATE 3. Initialize $\mathbf{W}^*$ by solving Eq.~(\ref{eq:plobj2}) with $\mathbf{Y}^* = \mathbf{Y}^{(s)}$;
\REPEAT
\STATE 4. SAP-I: find $\alpha^*$ with Eq.~(\ref{eq:apobj2}) and compute $\tilde{\mathbf{Y}}^*$ as $\tilde{\mathbf{Y}}^* = (\mathbf{V}_m \alpha^*)^T$;
\STATE 5. SAP-II: find $\mathbf{\bar{Y}}^*$ with Eq.~(\ref{eq:scobj2}) and compute $\mathbf{Y}^*$ as $\mathbf{Y}^* = \mathbf{\bar{Y}}^* + \mathbf{Y}^{(s)}$;
\STATE 6. BPL: find $\mathbf{W}^*$ by solving Eq.~(\ref{eq:plobj2});
\UNTIL{a stopping criterion is met}
\STATE 7. return $\mathbf{W}^*$.
\end{algorithmic}
\end{algorithm}

By joint SAP and BPL for inductive ZSL, our algorithm is given in Algorithm~\ref{alg:ours}. Once learned, given the optimal projection matrix $\mathbf{W}^*$ found by our algorithm, we predict the label of a test image $\mathbf{x}_i^{(u)}$ as:
\vspace{-0.05in}
\begin{small}
\begin{equation}
l_i^{(u)} = \arg\min_j \|\mathbf{x}_i^{(u)} - {\mathbf{W}^*}^T \mathbf{z}_j^{(u)}\|^2_2.
\end{equation}
\end{small}
\vspace{-0.15in}

Since each of iteration steps 4--6 in Algorithm~\ref{alg:ours} has an efficient solver and our algorithm is shown to converge very quickly ($\le 5$ iterations) in the experiments, it has a linear time complexity with respect to the data size.

\subsection{Algorithm Analysis}

We provide a rigorous analysis on the properties and behaviours of Algorithm~\ref{alg:ours} as follows. Without loss of generality, we first normalize all of $\|\mathbf{x}_i^{(s)}\|_2$, $\|\mathbf{y}_j^{(s)}\|_1$ to 1, and thus have: $\|\mathbf{Y}^{(s)}\|_F \le \|\mathbf{Y}^{(s)}\|_1 \le \sqrt{r}$.

\newtheorem{prop}{Proposition}
\begin{prop}
The solutions (i.e. $\mathbf{Y}^*$ and $\mathbf{W^*}$) found by Algorithm~\ref{alg:ours} are bounded.
\label{prop:bound}
\end{prop}

\begin{small}
\noindent\emph{Proof.} (a) Eq.~(\ref{eq:scobj2}) is equivalent to: $\mathbf{\bar{Y}}^* = \arg\min\limits_{\mathbf{\bar{Y}}} \mathrm{loss}(\mathbf{\bar{Y}})$, s.t. $\|\mathbf{\bar{Y}}\|_1 \leq M(\lambda_2)$, where $M(\lambda_2)$ is a constant depended on $\lambda_2$. Since $\mathbf{Y}^* = \mathbf{\bar{Y}}^* + \mathbf{Y}^{(s)}$, we have $\mathbf{\|{Y}^*\|}_F \le \mathbf{\|{\bar{Y}}^*\|}_F + \|{\mathbf{Y}^{(s)}\|}_F \le C_1$, where $C_1= M(\lambda_2) + \sqrt{r}$. \\
(b) Given that $\mathbf{Y}^*\mathbf{Y}^{*T}+\lambda_4\mathbf{I}$ and $\mathbf{X}^{(s)}\mathbf{X}^{(s)T}$ in Eq.~(\ref{eq:plobj2}) are nonnegative definite, there exist orthogonal matrices $\mathbf{P},\mathbf{Q}$ s.t. $\mathbf{\Sigma_1} \mathbf{P}^T\mathbf{WQ} + \mathbf{P}^T\mathbf{WQ}\mathbf{\Sigma_2} = 2\mathbf{P}^T\mathbf{Y}^*\mathbf{X}^{(s)T}\mathbf{Q}$, where $\mathbf{\Sigma_1} = \mathrm{diag}(\theta_1^{1},...\theta_k^{1})$ and $\mathbf{\Sigma_2} = \mathrm{diag}(\theta_1^{2},...\theta_d^{2})$ collect the eigenvalues of $\mathbf{Y}^*\mathbf{Y}^{*T}+\lambda_4\mathbf{I}$ and $\mathbf{X}^{(s)}\mathbf{X}^{(s)T}$, respectively. Obviously, $\theta_i^{1} \ge \lambda_4~(i=1,...,k), \theta_j^{2}\ge 0~(j=1,...,d).$ Let $\tilde{\mathbf{W}}=\mathbf{P}^T\mathbf{WQ}$ and $\mathbf{\tilde{R}} = \mathbf{P}^T\mathbf{Y}^*\mathbf{X}^{(s)T}\mathbf{Q}$. We have $\mathbf{\Sigma_1} \tilde{\mathbf{W}} + \tilde{\mathbf{W}}\mathbf{\Sigma_2} = 2 \mathbf{\tilde{R}} $. Since $\tilde{w}_{ij}=2\tilde{r}_{ij}/(\theta_i^{1} + \theta_j^{2})$, $\|\mathbf{W^*}\|_F = \|\tilde{\mathbf{W}}\|_F \le  2\|\mathbf{{Y}}^*\mathbf{X}^{(s)T}\|_F / \lambda_4$. Given that  $\mathbf{\|{Y}^*\|}_F  \le C_1$, we further obtain: $\|\mathbf{W^*}\|_F \le 2\|\mathbf{{Y}^*}\|_F \|\mathbf{X}^{(s)T}\|_F / \lambda_4 \le C_2$, where $C_2=2 C_1\sqrt{N_s}/\lambda_4$. \hfill $\square$
\end{small}

\begin{prop}
The optimal projection matrix $\mathbf{W}^*$ found by Algorithm~\ref{alg:ours} is insensitive to the perturbation of $\mathbf{Y}^*$, i.e., $\mathbf{\|{\triangle W}^*\|}_F \rightarrow 0$, if $\mathbf{\|{\triangle Y}^*\|}_F \rightarrow 0$.
\label{prop:robust}
\end{prop}

\begin{small}
\noindent\emph{Proof.} Given $\mathbf{W}^*$ found by Algorithm~\ref{alg:ours}, we have:
\begin{align}
(\mathbf{Y}^* \mathbf{Y}^{*T}+{\lambda_4}\mathbf{I}) \mathbf{W^*} + \mathbf{W^*} (\mathbf{X}^{(s)} \mathbf{X}^{(s)T}) = 2 \mathbf{Y}^* \mathbf{X}^{(s)T}.
\label{eq:proofw1}
\end{align}
When a perturbation ${\triangle \mathbf{Y}}^*$ is added to $\mathbf{Y}^*$, the optimal projection matrix found by Algorithm~\ref{alg:ours} is $\mathbf{\hat{W}}^*$:
\begin{align}
\mathbf{H}\mathbf{\hat{W}}^* + \mathbf{\hat{W}}^*(\mathbf{X}^{(s)} \mathbf{X}^{(s)T}) = 2(\mathbf{Y}^* +{\triangle \mathbf{Y}^*)}\mathbf{X}^{(s)T},
\label{eq:proofw2}
\end{align}
where $\mathbf{H} =  (\mathbf{{Y}}^* \hspace{-0.04in}+\hspace{-0.04in}  {\triangle \mathbf{Y}}^*)(\mathbf{Y}^* \hspace{-0.04in}+\hspace{-0.04in} {\triangle \mathbf{Y}^*})^{T}+{\lambda_4}\mathbf{I}$. Let $\triangle \mathbf{W}^* = \mathbf{\hat{W}}^* - \mathbf{W}^*$. Subtracting Eq.~(\ref{eq:proofw1}) from Eq.~(\ref{eq:proofw2}), we obtain:
\begin{align}
\mathbf{H}\triangle \mathbf{W}^* + \triangle \mathbf{W}^* (\mathbf{X}^{(s)} \mathbf{X}^{(s)T}) = \mathbf{K},
\end{align}
where $\mathbf{K} \hspace{-0.03in} = \hspace{-0.03in} 2 {\triangle \mathbf{Y}^*}\mathbf{X}^{(s)T} \hspace{-0.04in}-\hspace{-0.03in} ({\triangle \mathbf{Y}}^*{\triangle \mathbf{Y}}^{*T} \hspace{-0.04in}+\hspace{-0.04in} \mathbf{Y}^*{\triangle \mathbf{Y}}^{*T} \hspace{-0.04in}+\hspace{-0.04in} {\triangle \mathbf{Y}}^{*}\mathbf{Y}^{*T})$ $\mathbf{W}^*$. According to the proof of Prop.~\ref{prop:bound}, we similarly obtain: $\|\triangle \mathbf{W}^*\|_F \le \|\mathbf{K}\|_F / \lambda_4$. We further have that: $\|\triangle \mathbf{W}^*\|_F \le [2\sqrt{N_s}\|\triangle \mathbf{Y}^{*}\|_F + C_2\|\triangle \mathbf{Y}^{*}\|_F(\|\triangle \mathbf{Y}^{*}\|_F+ 2\|\mathbf{Y}^{*}\|_F)]/\lambda_4 \le {\|\triangle \mathbf{Y}^{*}\|_F[2\sqrt{N_s} + C_2(\|\triangle \mathbf{Y}^{*}\|_F + 2C_1)]}/\lambda_4$. This means that $\mathbf{\|{\triangle W}^*\|}_F \rightarrow 0$, if $\mathbf{\|{\triangle Y}^*\|}_F \rightarrow 0$.  \hfill $\square$
\end{small}

Note that Prop.~\ref{prop:bound} is used in the proof of Prop.~\ref{prop:robust} as a \emph{preliminary proposition}. Importantly, from Prop.~\ref{prop:robust}, the optimal projection matrix $\mathbf{W}^*$ used for final recognition is insensitive to the perturbation of $\mathbf{Y}^*$. This thus provides  guarantee that Algorithm~\ref{alg:ours} is robust under our new ZSL setting.

\subsection{ZSL with External Data}
\label{sect:exter}

Although the test images from unseen classes are not involved in the training process (see Algorithm~\ref{alg:ours}), the proposed algorithm still exploits the unannotated seen class images for SAP. Sometimes, even collecting unannotated seen class images becomes a burden. To address this issue, we thus choose to perform SAP with the unannotated external data from image search engine. By searching relevant images with high-level semantic abstraction (i.e. query) of seen classes, we obtain many free web images to augment the few annotated seen class images at hand. These unannotated web images can be readily exploited for SAP, instead of the unannotated seen class images used in Algorithm~\ref{alg:ours}. When the few annotated seen class images are fused with the unannotated external data, the proposed algorithm can be implemented without any modifications.

Note that the unannotated web images are obtained at a low cost (search key words on a image search engine), and thus unavoidably contain some images that do not belong to the seen classes.  For example, given the benchmark seen/unseen class split (i.e. 150/50) of the CUB-200-2011 Birds (CUB) dataset \cite{CUB-200-2011}, we collect the external data by Google with the query `North American Bird' (i.e. high-level semantic abstraction of seen classes). With this high-level query, it is very likely that a returned image comes from either seen or unseen classes, and beyond (see Figure~\ref{fig:images}). In this paper, we choose to classify the obtained web images using the CNN model proposed in \cite{yu2018hierarchical}, and then discard the images that are classified to unseen classes. Given that \cite{yu2018hierarchical} has reported a very high accuracy in fine-grained classification, \emph{the effect of possible unseen class images can be suppressed dramatically} during training our SAP model. Therefore, the achieved improvements (if any) are mainly contributed to our SAP model itself (trained with a few annotated seen class images and a number of unannotated web images mostly from seen classes).

\vspace{0.2cm}
\section{Generalization to Social Image Annotation}

Since the noisy tags of social images are analogous to the attributes in ZSL, our model can be naturally applied to social image annotation (SIA). Concretely, we first replace $\mathbf{Y}^{(s)}$ with the semantic representation defined with noisy tags, and then refine $\mathbf{Y}^{(s)}$ by running our algorithm. By concatenating $\mathbf{Y}^*$ and $\mathbf{X}^{(s)}$, we finally train multi-class classifiers (i.e. one-vs-all SVM) for SIA. Note that it is still time consuming to find the $m$ smallest eigenvectors of $\mathbf{L}$ on an extremely large dataset (e.g. NUS-WIDE \cite{chua2009civr}). To keep the scalability of our model, we thus employ nonlinear approximation to find $m$ smallest eigenvectors as in \cite{chen2011large}. In this application, our model can be fairly compared to the state-of-the-art alternatives \cite{wang2016cvpr,jin2016icpr,hu2016cvpr,liu2017semantic}.

\begin{table}[t]
\begin{center}
\begin{footnotesize}
\tabcolsep0.2cm
\begin{tabular}{c|c|c|c}
\hline
Dataset & \# images & \# attributes & \# seen/unseen classes\\
\hline
AwA & 30,475 & 85 & 40/10 \\
CUB & 11,788 & 312 & 150/50 \\
aPY & 15,339 & 64 & 20/12 \\
SUN & 14,340 & 102 & 707/10 \\
\hline
\end{tabular}
\end{footnotesize}
\end{center}
\vspace{-0.05in}
\caption{Details of the four benchmark datasets.}
\label{tab:datasets}
\vspace{-0.00in}
\end{table}

\vspace{-0.1cm}
\section{Experiments}

\subsection{ZSL on Benchmark Datasets}
\label{sect:zslbench}

\subsubsection{Datasets and Settings}
\vspace{-0.1cm}

\noindent\textbf{1) Datasets}. Four widely-used benchmark datasets are selected: Animals with Attributes (AwA) \cite{Lampert2014pami}, CUB-200-2011 Birds (CUB) \cite{CUB-200-2011}, aPascal\&Yahoo (aPY) \cite{Farhadi2009cvpr}, and SUN Attribute (SUN) \cite{Patterson2014ijcv}. The details of these benchmark datasets are given in Table~\ref{tab:datasets}.

\noindent\textbf{2) Semantic and Feature Spaces}. First, we establish the semantic space with attributes for the four benchmark datasets, all of which provide the attribute annotations for seen/unseen classes. Second, we extract the Resnet101 \cite{he2016cvpr} features to form the visual feature space as in \cite{Chen2018CVPR,Sung2018CVPR,Xian2018CVPR}, which has been widely used by many existing ZSL models.

\noindent\textbf{3) Evaluation Metrics}. For the standard ZSL setting, we compute the multi-way classification accuracy as in previous works. For the generalized ZSL setting \cite{chao2016empirical,rahman2017unified,Xian2017CVPR}, we compute the harmonic mean of two accuracies to evaluate the overall performance: $acc_u$ -- the accuracy of classifying the test samples from unseen classes to all seen/unseen classes, and $acc_s$ -- the accuracy of classifying the test samples from seen classes to all seen/unseen classes.

\noindent\textbf{4) Parameter Settings}. Our algorithm has five parameters to tune: $k_g, m, \lambda_1, \lambda_2, \lambda_3$. Given only few annotated seen class images, it is impossible to select the parameters by cross-validation. Fortunately, our algorithm is shown to be insensitive to these parameters (see the supporting figures in the suppl. material), and thus we uniformly set $k_g=300$, $m=50$, $\lambda_1=0.01$, $\lambda_2=1e-4$, and $\lambda_3=1e-6$.

\noindent\textbf{5) Compared Methods}. We make comparison to five representative ZSL models: RPL \cite{Shigeto2015}, ESZSL \cite{Romera2015icml}, SSE \cite{Zhang2015iccv}, ZSKL \cite{ZK2018CVPR}, and RN \cite{Sung2018CVPR}. The baselines are selected based on two criteria: (1) Latest/state-of-the-art (e.g. the CVPR'18 papers \cite{ZK2018CVPR,Sung2018CVPR}). (2) For comparison in Table~\ref{tab:external}, some baselines are selected because they can utilize the propagated attributes (with continuous, rather than binary  values) as inputs for ZSL.

\begin{table}[t]
\begin{center}
\begin{footnotesize}
\tabcolsep0.05cm
\begin{tabular}{c|c|c|c|c|c|c|c}
\hline
Dataset & $K$ & RPL &  ESZSL & SSE  & ZSKL & RN & Ours \\
\hline
& 5 & 29.4(1.4) & 24.1(1.4) & 15.6(2.8) & 32.2(1.6) & 27.5(7.9) & \textbf{40.9}(\textbf{1.5}) \\
& 10 & 34.3(1.0) & 27.7(0.8) & 16.3(1.4) & 37.8(0.8) & 28.0(3.2) &  \textbf{45.8}(\textbf{0.8})\\
CUB & 15 & 38.2(0.9) & 30.3(0.6) & 18.2(0.9) & 40.5(0.5) & 31.3(6.5) &  \textbf{47.5}(\textbf{0.8})\\
& 20 & 40.0(0.5) & 32.9(0.4) & 19.2(1.2) & 41.5(0.7) & 32.6(4.0) &  \textbf{48.4}(\textbf{0.3})\\
& 25 & 41.4(1.0) & 34.9(0.6) & 21.0(0.6) & 42.3(0.3) & 35.4(1.8) &  \textbf{49.3}(\textbf{0.7})\\
\hline
& 5 & 50.3(2.0) & 26.4(4.0) & 39.9(3.2) & 54.5(2.7) & 28.7(3.2) & \textbf{71.0}(\textbf{3.1})\\
& 10 & 53.6(2.2) & 26.8(5.7) & 41.4(5.8) & 61.0(2.2) & 29.7(1.8) & \textbf{74.5}(\textbf{1.8})\\
AwA& 15 & 53.8(2.3) & 34.4(1.8) & 42.2(4.4) & 62.6(2.4) & 31.1(4.0) & \textbf{75.7}(\textbf{1.7})\\
& 20 & 55.2(1.7) & 40.8(1.7) & 42.6(4.5) & 65.6(1.5) & 32.4(5.2) & \textbf{76.0}(\textbf{1.4})\\
& 25 & 55.7(1.4) & 41.5(2.0) & 42.9(3.5) & 66.7(1.9) & 35.2(3.3) &  \textbf{77.5}(\textbf{0.9})\\
\hline
& 5 & 21.4(5.7) & 19.2(4.4) & 13.1(3.0) & 33.6(3.4) & 8.0(4.3) & \textbf{42.5}(\textbf{8.3})\\
& 10 & 22.3(3.6) & 19.8(4.0) & 14.0(3.4) & 35.0(6.1) & 27.4(4.6) & \textbf{44.1}(\textbf{4.8})\\
aPY & 15 & 23.0(2.6) & 20.8(5.3) & 14.1(4.0) & 37.1(4.7) & 32.2(2.8) & \textbf{44.8}(\textbf{5.0})\\
& 20 & 24.9(2.8) & 20.6(3.4) & 15.3(2.3) & 38.7(6.5) & 32.7(2.8) & \textbf{45.7}(\textbf{4.7})\\
& 25 & 25.5(2.8) & 21.8(2.1) & 17.6(2.1) & 40.4(4.6) & 35.1(2.5) & \textbf{47.4}(\textbf{4.7})\\
\hline
& 1 & 57.2(3.2) & 58.0(4.4) & 58.1(3.1) & 58.9(5.5) &  54.2(3.8) & \textbf{81.7}(\textbf{1.9})\\
& 2 & 62.4(3.3) & 62.5(4.7) & 60.2(3.3) & 67.8(1.6) & 58.7(4.8) & \textbf{83.0}(\textbf{2.2})\\
SUN & 3 & 64.0(4.1) & 65.8(5.2) & 60.8(3.2) & 70.3(2.4) & 60.4(4.0) &  \textbf{83.3}(\textbf{1.4})\\
& 4 & 66.5(3.2) & 68.9(4.5) & 62.1(3.4) & 71.4(2.6) & 62.1(5.5) &  \textbf{83.9}(\textbf{1.4})\\
& 5 & 69.1(1.9) & 70.2(2.1) & 62.6(2.2) & 73.4(2.2) & 64.6(4.4) &  \textbf{84.3}(\textbf{1.3})\\
\hline
\end{tabular}
\end{footnotesize}
\end{center}
\vspace{-0.05in}
\caption{Comparative results (\%) of standard ZSL. Average accuracy is reported along with standard deviation (in brackets).}
\label{tab:zslmed}
\vspace{-0.12in}
\end{table}

\vspace{-0.3cm}
\subsubsection{Results of Standard ZSL}
\vspace{-0.2cm}

The comparative results under the standard ZSL setting are shown in Table~\ref{tab:zslmed}. Note that all seen class images from each dataset are provided for training, but only $K$ images per seen class are annotated (the other are unannotated). For \emph{fair comparison}, all five ZSL alternatives apply the nearest neighbor classifier over few annotated seen class images to classify each unannotated image to a seen class (thus its attribute vector can be obtained). We have the following observations: (1) Our model achieves the best performance on all four datasets, and the improvements over the second-best competitor range from 7\% to 23\%. This clearly validates the effectiveness of our model in overcoming the attribute sparsity by integrating SAP and BPL for inductive ZSL. (2) The performance margin between our model and five ZSL alternatives generally becomes bigger as fewer annotated seen class images are provided for model training. Our explanation is that: other than five ZSL alternatives, our model exploits more accurately propagated attribute annotations (obtained by SAP) for BPL. (3) Our model significantly outperforms the state-of-the-art deep RN model \cite{Sung2018CVPR}, which is really impressive since deep learning is not considered in our model. This also suggests that a deep ZSL model tends to suffer from the annotation sparsity and thus may be not suitable for our new ZSL setting defined in this paper.

\vspace{-0.2cm}
\subsubsection{Results of Generalized ZSL}
\vspace{-0.1cm}

We follow the same generalized ZSL setting of \cite{chao2016empirical}, i.e., 20\% of the images from the seen classes are held out and then mixed with the images from the unseen classes to form the test set. We provide $K$ annotated images per seen class for model training as in the standard ZSL setting. The comparative results on the AwA and CUB datasets are presented in Table~\ref{tab:gzsl}. Our model is still shown to achieve the best performance under this more challenging ZSL setting. Importantly, the obtained even bigger margins suggest that both SAP and BPL can promote the generalization ability of our model for generalized ZSL.

\begin{table}[t]
\begin{center}
\begin{footnotesize}
\tabcolsep0.05cm
\begin{tabular}{c|c|c|c|c|c|c|c}
\hline
Dataset & $K$ & RPL &  ESZSL & SSE  & ZSKL & RN & Ours \\
\hline
& 5 & 16.8(1.5) & 9.4(0.6) & 6.9(1.4) & 18.6(0.9) & 12.4(2.5) & \textbf{27.7}(\textbf{0.5}) \\
& 10 & 21.3(0.8) & 9.9(0.5) & 8.7(1.4) & 22.2(0.5) & 14.3(1.4) & \textbf{33.9}(\textbf{0.4}) \\
CUB & 15 & 23.7(0.3) & 10.6(0.9) & 9.0(0.6) & 24.1(0.3) & 15.3(2.6) & \textbf{36.1}(\textbf{0.9}) \\
&  20 & 25.3(0.2) & 11.8(0.4) & 9.9(0.3) & 25.3(0.3) & 17.3(1.4) & \textbf{37.9}(\textbf{0.7}) \\
&  25 & 26.1(0.3) & 12.7(1.4) & 10.5(1.2) & 26.0(0.3) & 19.9(4.6) & \textbf{39.0}(\textbf{0.3}) \\
\hline
& 5 & 40.6(1.6) & 27.7(4.1) & 19.4(2.4) & 45.6(1.7) & 36.8(2.4) &   \textbf{55.1}(\textbf{2.8}) \\
& 10 & 41.3(1.7) & 29.2(4.0) & 21.7(2.2) & 46.9(1.7) & 37.8(4.8) & \textbf{59.0}(\textbf{1.7})\\
AwA &  15 & 41.4(0.9) & 33.2(2.6) & 24.2(2.6) & 47.4(1.8) & 40.3(1.8) &  \textbf{61.2}(\textbf{1.3})\\
&  20 & 41.5(1.6) & 37.1(1.5) & 28.6(0.6) & 48.1(1.5) & 41.7(0.6) &  \textbf{62.7}(\textbf{1.3})\\
&  25 & 41.6(1.4) & 39.7(2.4) & 32.5(0.6) & 48.5(1.5) & 42.7(3.5) &  \textbf{63.0}(\textbf{2.0})\\
\hline
\end{tabular}
\end{footnotesize}
\end{center}
\vspace{-0.05in}
\caption{Comparative results (\%) of generalized ZSL. Harmonic mean is reported along with standard deviation (in brackets).}
\label{tab:gzsl}
\vspace{-0.00in}
\end{table}

\begin{figure}[t]
\vspace{0.05in}
\begin{center}
\includegraphics[width=0.48\textwidth]{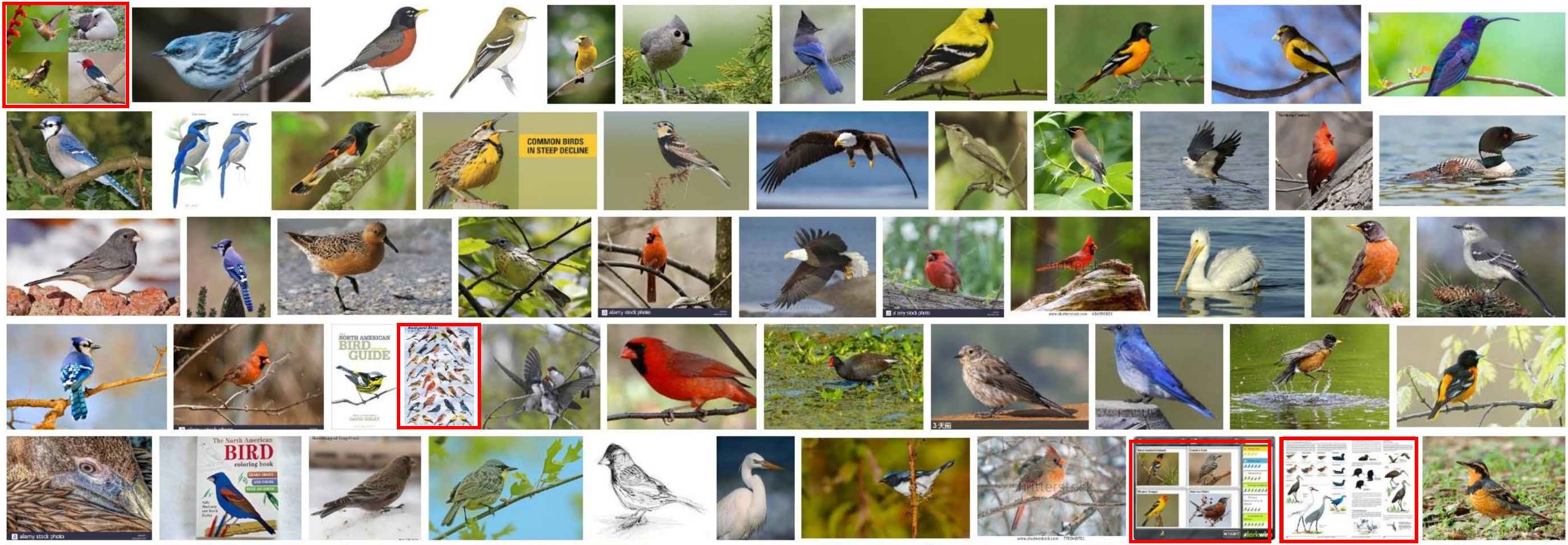}
\end{center}
\vspace{-0.1in}
\caption{Examples of the top images returned by Google with the query `North American Bird'. We directly discard the images with bird objects from multiple classes (marked with red boxes). } \label{fig:images}
\vspace{-0.05in}
\end{figure}

\vspace{-0.0cm}
\subsection{ZSL with External Data}

\vspace{-0.0cm}
\subsubsection{Dataset and Settings}

We construct a new dataset called CUB+Web\footnote{\url{https://github.com/anonymous04321/cub-web}} as follows: 1) The training set has 750 annotated images (5 per class) from the 150 seen classes of standard CUB \cite{CUB-200-2011}, along with 1,205 unannotated web images obtained by Google with the query `North American Bird'; 2) The test set has 2,946 unannotated images from the 50 unseen classes of CUB. Particularly, we first download top-2,000 web images from Google and discard the images with bird objects from multiple classes (see Figure~\ref{fig:images}). Furthermore, we classify the obtained web images using the CNN model proposed in \cite{yu2018hierarchical}, and discard the images that are classified to unseen classes, resulting in 1,205 unannotated web images left. Since \cite{yu2018hierarchical} has reported a very high accuracy in fine-grained classification, \emph{the effect of possible unseen class images can be suppressed dramatically} during training our SAP model. Additionally, for CUB+Web, the semantic and feature spaces are formed exactly the same as in Sec.~\ref{sect:zslbench}.

\vspace{-0.1cm}
\subsubsection{Comparative Results}

We make comparison to six closely related ZSL models: 1) RPL0 -- the reverse projection learning model \cite{Shigeto2015} trained only with few annotated seen class images; 2) RPL -- the RPL model \cite{Shigeto2015} trained with not only few annotated seen class images but also the web images with propagated attributes obtained by our model; 3) ESZSL0 -- the ESZSL model \cite{Romera2015icml} trained only with few annotated seen class images; 4) ESZSL -- the ESZSL model \cite{Romera2015icml} trained like RPL (but not RPL0); 5) ZSKL0 -- the ZSKL model \cite{ZK2018CVPR} trained only with few annotated seen class images; 6) ZSKL -- the ZSKL model \cite{ZK2018CVPR} trained like RPL (but not RPL0). Since both SSE \cite{Zhang2015iccv} and RN \cite{Sung2018CVPR} take the class labels of training images as inputs, they are not applied to ZSL with external data (having no exact class labels).

\begin{table}[t]
\begin{center}
\begin{footnotesize}
\tabcolsep0.17cm
\begin{tabular}{c|c|c|c|c|c|c|c}
\hline
$K$ & RPL0 & RPL & ESZSL0 & ESZSL & ZSKL0 & ZSKL &  Ours \\
\hline
1 & 21.5 & 26.9 & 10.4 & 12.8 & 23.0 & 26.5 & \textbf{29.1 }\\
2 & 23.5 & 29.4 & 20.6 & 23.5 & 26.2 & 30.3 & \textbf{32.9 }\\
3 & 26.2 & 31.4 & 22.1 & 27.6 & 28.5 & 32.7 & \textbf{36.4 }\\
4 & 28.9 & 35.0 & 23.5 & 30.8 & 30.5 & 34.5 & \textbf{39.1 }\\
5 & 29.4 & 37.3 & 24.1 & 32.2 & 32.2 & 35.8 & \textbf{41.2 }\\
\hline
\end{tabular}
\end{footnotesize}
\end{center}
\vspace{-0.05in}
\caption{Comparative accuracies (\%) of ZSL with external data. }
\label{tab:external}
\vspace{0.05in}
\end{table}

\begin{figure}[t]
\vspace{0.00in}
\begin{center}
\includegraphics[width=0.98\columnwidth]{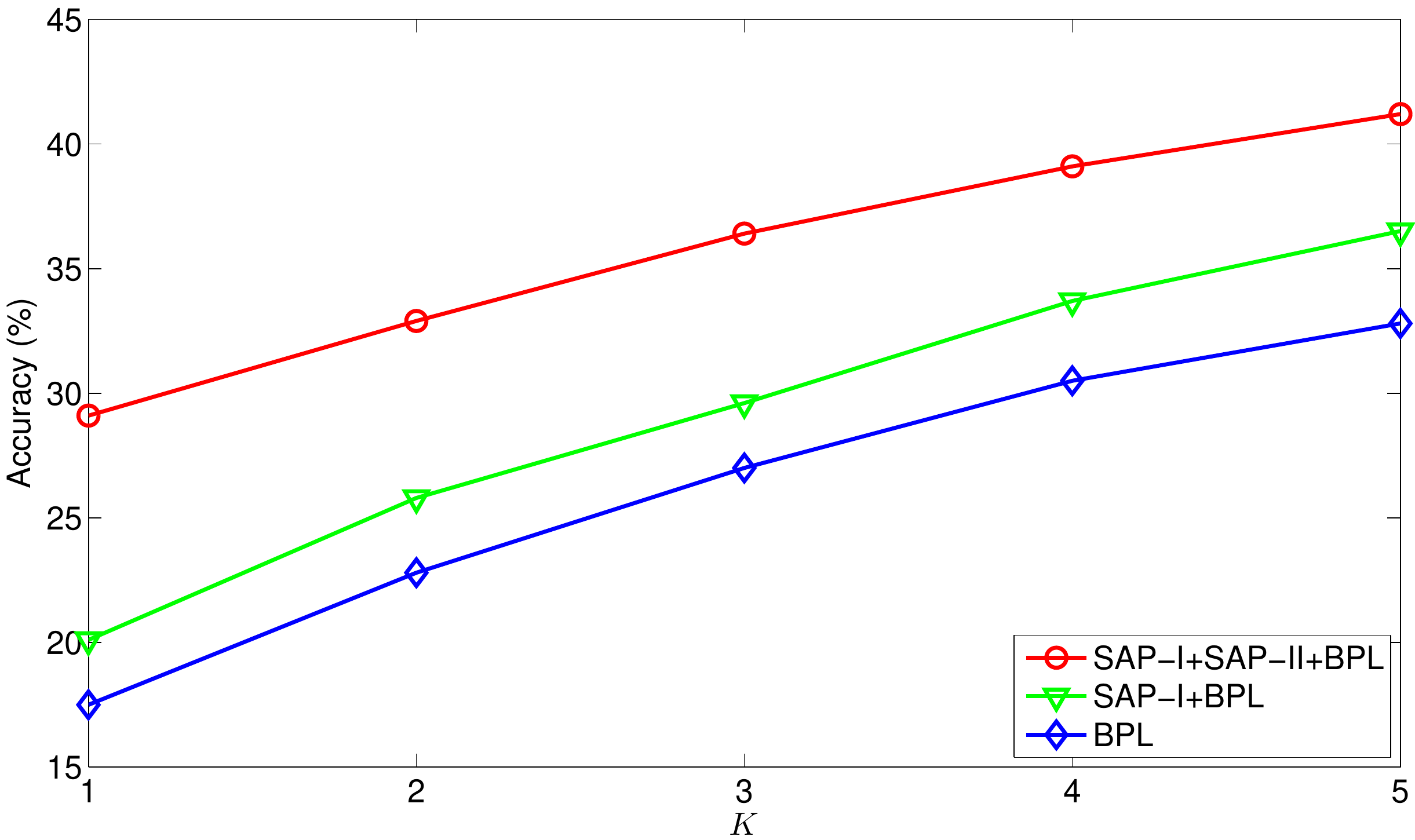}
\end{center}
\vspace{-0.15in}
\caption{Ablation study results on the CUB+Web dataset. } \label{fig:comp}
\vspace{-0.05in}
\end{figure}

The comparative results are presented in Table~\ref{tab:external}, where $K$ annotated images per seen class are provided for model training. It can be seen that: (1) The four models (i.e. RPL, ESZSL, ZSKL, and ours) trained using extra web images with propagated attributes lead to significant improvements over those without using extra web images (i.e. RPL0, ESZSL0, and ZSKL0), validating the effectiveness of our sparse attribute propagation. (2) Due to bidirectional projection learning, our model achieves better results than RPL, ESZSL, and ZSKL under exactly the same setting.

\begin{figure}[t]
\vspace{0.05in}
\begin{center}
\includegraphics[width=0.97\columnwidth]{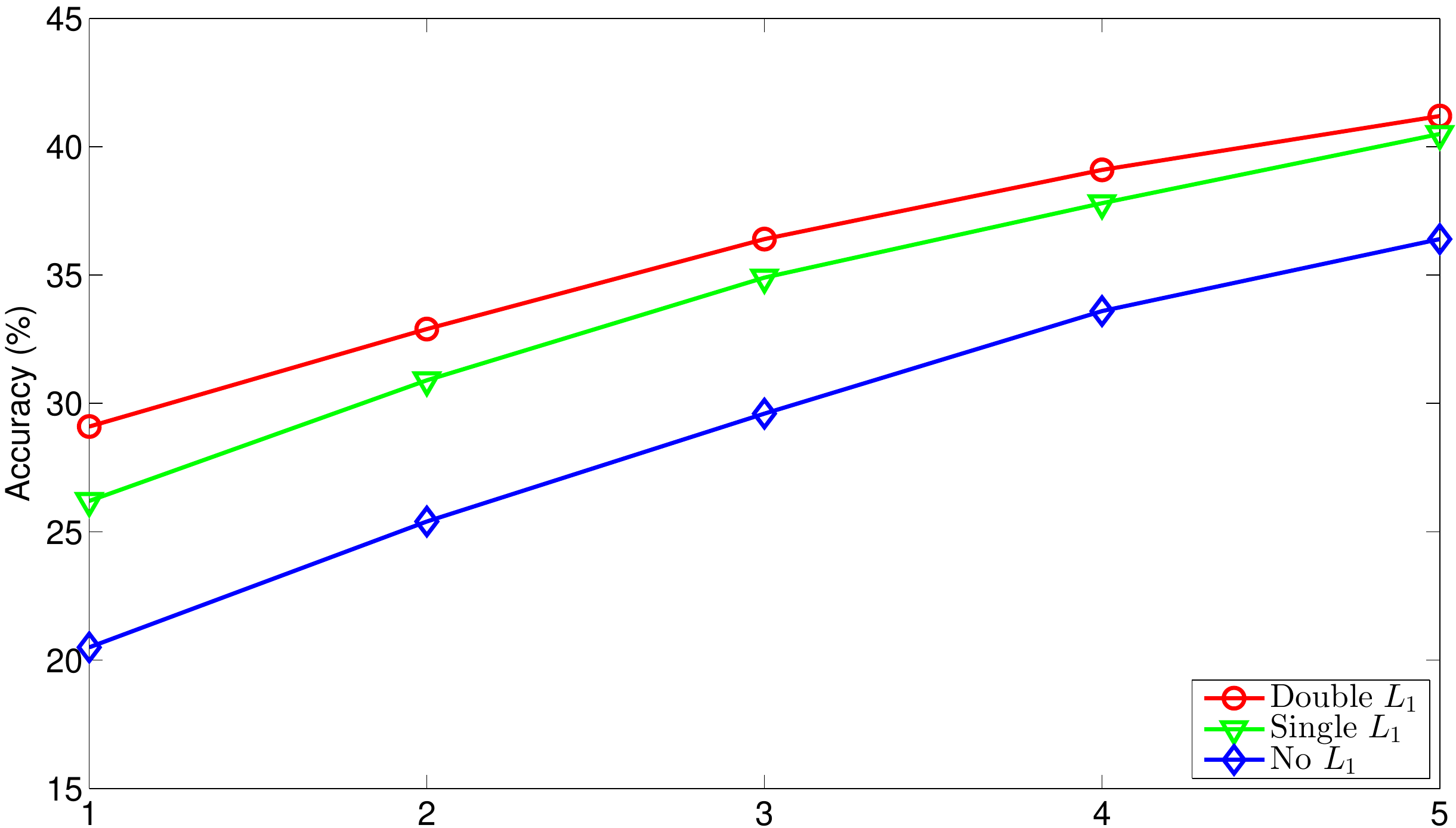}
\end{center}
\vspace{-0.12in}
\caption{Comparative results obtained by different attribute propagation models on the CUB+Web dataset. } \label{fig:sparse}
\vspace{-0.12in}
\end{figure}

\begin{table*}[t]
\vspace{0.05in}
\centering
\begin{center}
\begin{footnotesize}
\tabcolsep0.3cm
\begin{tabular}{l|c|ccc|ccc}
\hline
Methods   & Label correlation & C-P (\%)  & C-R (\%) & C-F1 (\%)  & I-P (\%)  & I-R (\%)  & I-F1 (\%) \\
\hline
Ours (Refined Tags+CNN+SVM) & no &64.79 & \bf68.16 & \bf66.43 & \bf77.71 & 70.69 & 74.03 \\
Ours (Refined Tags+SVM) & no &56.70 & 58.79 & 57.73 & 70.65 & 61.31 & 65.65 \\
Ours (Noisy Tags+SVM) & no &14.13 & 53.75 & 22.38 & 22.75 & 47.91 & 30.85 \\
\hline
SR-CNN-RNN \cite{liu2017semantic}  & yes & \bf71.73 &61.73 & 66.36 & 77.41 & 76.88 & \bf77.15  \\
SINN \cite{hu2016cvpr}    & yes & 58.30 & 60.30 & 59.44 & 57.05 & \bf79.12 & 66.30 \\
TagNeighboor \cite{johnson2015iccv} & no & 54.74 & 57.30 & 55.99 & 53.46 & 75.10 & 62.46 \\
RIA   \cite{jin2016icpr}  & yes & 52.92 & 43.62 & 47.82 & 68.98 & 66.75 & 67.85   \\
CNN-RNN  \cite{wang2016cvpr}  & yes  & 40.50 & 30.40 & 34.70 & 49.90 & 61.70 & 55.20  \\
CNN+WARP \cite{gong2013arxiv}   & no & 31.65 & 35.60 & 33.51 & 48.59 & 60.49 & 53.89   \\
CNN+softmax \cite{gong2013arxiv}  & no & 31.68 & 31.22 & 31.45 & 47.82 & 59.52 & 53.03\\
CNN+logistic \cite{hu2016cvpr}  & no & 45.60 & 45.03 & 45.31 & 51.32 & 70.77 & 59.50  \\
\hline
\end{tabular}
\end{footnotesize}
\end{center}
\vspace{-0.05in}
\caption{Comparative results of social image annotation on the NUS-WIDE dataset. }
\label{tab:sia}
\vspace{-0.15in}
\end{table*}

\vspace{-0.1cm}
\subsubsection{Further Evaluation}

\noindent\textbf{Ablation Study.} To evaluate the contribution of each component (SAP-I, SAP-II, or BPL), we conduct experiments by adding more components to the BPL model. The ablation study results in Figure~\ref{fig:comp} show that: (1) The SAP-I step by solving Eq.~(\ref{eq:apobj2}) yields better results in all cases (see SAP-I+BPL vs.~BPL). (2) The SAP-II step by solving Eq.~(\ref{eq:scobj2}) further leads to performance improvements (see SAP-I+SAP-II+BPL vs.~SAP-I+BPL), which become more significant with fewer annotated seen class images (i.e. smaller $K$).

\noindent\textbf{Alternative Attribute Propagation.} We compare three alternative attribute propagation models: 1) `Double $L_1$': our model formulated in Eq.~(\ref{eq:obj}); 2) `Single $L_1$': $\|\mathbf{B} \tilde{\mathbf{Y}}^T\|_1$ used in our model is replaced by $\|\mathbf{B} \tilde{\mathbf{Y}}^T\|_F^2$; 3) `No $L_1$': $\|\mathbf{Y} \hspace{-0.05in}-\hspace{-0.04in} \mathbf{Y}^{(s)}\|_1$ is removed from the second model `Single $L_1$'. The comparative results are presented in Figure~\ref{fig:sparse}. As expected, more $L_1$-norm regularization terms used for attribute propagation lead to better results.

\vspace{-0.0cm}
\subsection{Social Image Annotation}

\vspace{-0.1cm}
\subsubsection{Dataset and Settings}
\vspace{-0.2cm}

We make performance evaluation on the NUS-WIDE \cite{chua2009civr} benchmark dataset, which consists of 269,648 images and 81 class labels from Flickr image metadata. By removing invalid Flickr links and also images with no social tags, we obtain 162,806 training images and 57,202 test images. In this paper, we keep 1,000 most frequent tags to form the semantic space, and finetune ResNet101 \cite{he2016cvpr} with the training set to extract the visual features.

The per-class and per-image metrics including precision and recall have been widely used in previous works. In the following, the per-class precision (C-P) and per-class recall (C-R) are obtained by computing the mean precision and recall over all the classes, while the overall per-image precision (I-P) and overall per-image recall (I-R) are computed by averaging over all the test images. Moreover, the per-class F1-score (C-F1) and overall per-image F1-score (I-F1) are used for comprehensive performance evaluation by combining precision and recall with the harmonic mean.

\vspace{-0.2cm}
\subsubsection{Comparative Results}
\vspace{-0.2cm}

Our model has three variants: 1) Ours (Noisy Tags+SVM) -- SVM trained with the original noisy tags; 2) Ours (Refined Tags+SVM) -- SVM trained with the refined tags; 3) Ours (Refined Tags+CNN+SVM) -- SVM trained with the refined tags and CNN features. Moreover, we also make comparison to the state-of-the-art models \cite{liu2017semantic,hu2016cvpr,johnson2015iccv,jin2016icpr,wang2016cvpr,gong2013arxiv}. The comparative results in Table~\ref{tab:sia} show that: (1) The refined tags obtained by our model yield over 30\% improvements when C-F1 and I-F1 are concerned. (2) Our model achieves state-of-the-art performance according to C-F1 and competitive results according to I-F1. This is really impressive given that the label correlation is not exploited for social image annotation in our model.

\vspace{-0.1cm}
\section{Conclusion}

In this paper, we have investigated the challenging problem of ZSL with less human annotation. For the first time, we have defined a new ZSL setting where only few annotated seen class images are given for training. To overcome the annotation sparsity, we have proposed a novel inductive ZSL model by formulating SAP and BPL within a unified framework, with rigorous theoretic algorithm analysis provided. Moreover, we have generalized the proposed model to ZSL with external data and also social image annotation. Extensive experiments have shown that the proposed model achieves state-of-the-art performance.

\end{document}